\documentclass[runningheads]{llncs}
\pdfoutput=1

\usepackage{subfigure}
\usepackage[hidelinks]{hyperref}
\usepackage{url}
\usepackage{graphicx}
\usepackage{amsfonts}
\usepackage{amsmath}
\usepackage{multirow}
\usepackage{pgfplots}
\usepackage{pgfplotstable} 
\pgfplotsset{compat=newest}
\usepackage{enumitem}
\usepackage{wrapfig}
\usepackage{flushend}
\usepackage{arydshln}
\usepackage{microtype}
\usepackage{todonotes}

\newcommand{\citet}{\cite}
\newcommand{\citep}{\cite}

\title{Two-view Graph Neural Networks \\ for Knowledge Graph Completion}

\author{Vinh Tong$^1$, Dai Quoc Nguyen$^2$, Dinh Phung$^3$, Dat Quoc Nguyen$^4$}

\authorrunning{Tong et al.}

\institute{$^1$University of Stuttgart, $^2$Oracle Labs, $^3$Monash University, $^4$VinAI Research\\
\email{$^1$vinh.tong@ipvs.uni-stuttgart.de, $^2$dai.nguyen@oracle.com,}\\
\email{$^3$dinh.phung@monash.edu, $^4$v.datnq9@vinai.io}
}

\begin{document}

\maketitle

\begin{abstract}
We present an effective graph neural network (GNN)-based knowledge graph embedding model, which we name WGE, to capture entity- and relation-focused graph structures. Given a knowledge graph, WGE builds a single undirected entity-focused graph that views entities as nodes. WGE also constructs another single undirected graph from relation-focused constraints, which views entities and relations as nodes. WGE then proposes a GNN-based architecture to better learn vector representations of entities and relations from these two single entity- and relation-focused graphs. WGE feeds the learned entity and relation representations into a weighted score function to return the triple scores for knowledge graph completion. Experimental results show that WGE outperforms strong baselines on seven benchmark datasets for knowledge graph completion.


\end{abstract}

\section{Introduction}\label{sec:intro}

A knowledge graph (KG) is a network of entity nodes and relationship edges, which can be represented as a collection of triples in the form of \textit{(h, r, t)}, wherein each triple \textit{(h, r, t)} represents a relation $r$ between a head entity $h$ and a tail entity $t$. Here, entities are real-world things or objects such as  music tracks, movies persons, organizations, places and the like, while each relation type determines a certain relationship between entities. KGs  are used in many commercial applications, e.g. in  such search engines as Google, Microsoft's Bing and Facebook's Graph search. They also  are useful resources for many natural language processing tasks such as co-reference resolution \citep{ponzetto-strube:2006:HLT-NAACL06-Main,TACL522},  semantic parsing \citep{krishnamurthy-mitchell:2012:EMNLP-CoNLL,berant-EtAl:2013:EMNLP} and question answering  \citep{Ferrucci:2012:ITW:2481742.2481743,Fader:2014:OQA:2623330.2623677}. However, an issue is that KGs are often incomplete, i.e., missing a lot of valid triples \citep{bordes2011learning,Nguyen2020KGC}. For an example of a specific application, question answering systems based on incomplete KGs would not provide correct answers given correctly interpreted input queries. 
Thus, much work has been devoted towards KG completion to perform link prediction in KGs. In particular, many KG embedding models have been proposed to predict whether a triple not in KGs is likely to be valid or not, e.g., TransE \citep{NIPS2013_5071}, DistMult \citep{Yang2015}, ComplEx \citep{Trouillon2016} and QuatE \citep{zhang2019quaternion}.
These KG embedding models aim to learn vector representations for entities and relations and define a score function such that \textit{valid triples have higher scores than invalid ones} \citep{Nguyen2020KGC,zhang2019autosf}, e.g.,  the score of the valid triple (Sydney, city\_in, Australia) is higher than the score of the invalid one (Sydney, city\_in, Vietnam).

Recently, several KG completion works have adapted graph neural networks (GNNs) using an encoder-decoder architecture, e.g., R-GCN \citep{schlichtkrull2017modeling} and CompGCN \citep{vashishth2020compositionbased}.
In general, the encoder module customizes GNNs to update vector representations of entities and relations. Then, the decoder module employs an existing score function  to return the triple score \citep{NIPS2013_5071,Yang2015,Trouillon2016,Dettmers2017,Nguyen2018ConvKBfull,978-3-030-77385-4_24,pmlr-v180-chen22c}. 
For example, R-GCN adapts Graph Convolutional Networks (GCNs) \citep{kipf2017semi} to construct a specific encoder to update only entity embeddings. 
CompGCN modifies GCNs to use composition operations between entities and relations in the encoder module.
Note that these existing GNN-based KG embedding models mainly consider capturing the graph structure surrounding entities as relation representations are used to update the entity embeddings only (as shown in Equations \ref{eqn:grcn}, \ref{eqn:compgcn} and \ref{eqn:compgcn2}; and see the last paragraph of Section \ref{sec:relatedwork} for a detailed discussion). Therefore, they might miss covering potentially useful information on relation structure. 
 
To this end, we propose a new KG embedding model---named WGE that is equivalent to VVGE to abbreviate Two-View Graph Embedding---to leverage GNNs to capture both entity-focused graph structure and relation-focused graph structure for KG completion.
In particular, WGE transforms a given KG into two views. 
The first view---a single undirected entity-focused graph---only includes entities as nodes to provide the entity neighborhood information. 
The second view---a single undirected relation-focused graph---considers both entities and relations as nodes, constructed from constraints (\textit{subjective relation, predicate entity, objective relation}) e.g. (born\_in, Sydney, city\_in), to attain the potential dependence between two neighborhood relations.  For instance, the knowledge about a potential dependence between ``born\_in'' and ``city\_in'' could be relevant for predicting some other relationship, e.g. ``nationality'' or  ``country of citizenship''. Then WGE introduces a new GNN-based encoder module that directly takes these two graph views as input to better update entity and relation embeddings. WGE feeds the entity and relation embeddings into its decoder module that uses a weighted score function to return the triple scores for KG completion. 
In summary, our contributions are as follows: 

\begin{itemize}
    \item We present WGE for KG completion, that first proposes to transform a given KG into entity- and relation-focused graph structures and then introduces a new encoder architecture to learn entity and relation embeddings from these two graph structures. 

    \item To verify model effectiveness, we conduct extensive experiments to compare our WGE with other strong GNN-based baselines on seven benchmark datasets, including FB15K-237 \citep{toutanova-chen-2015-observed} and six new and difficult datasets of CoDEx-S, CoDEx-M, CoDEx-L, LitWD1K, LitWD19K and LitWD48K \citep{safavi2020codex,LiterallyWikidata}.  The experiments show that WGE outperforms the GNN-based baselines and other competitive KG embedding models on these seven datasets. 
    
\end{itemize}

\section{Related work} 
\label{sec:relatedwork}

Recently, GNNs become a central strand to learn low-dimensional continuous embeddings for nodes and graphs \citep{scarselli2009graph,hamilton2017representation}.
GNNs provide faster and more practical training and state-of-the-art results on benchmark datasets for downstream tasks \citep{wu2019comprehensive,zhang2020network}.
In general, GNNs update the vector representation of each node by transforming and aggregating the vector representations of its neighbors \citep{kipf2017semi,hamilton2017inductive,velickovic2018graph,Nguyen2020QGNN,NoGE}.

We represent each graph $\mathcal{G} = \left(\mathcal{V}, \mathcal{E}\right)$, where $\mathcal{V}$ is a set of nodes; and $\mathcal{E}$ is a set of edges. 
Given a graph $\mathcal{G}$, we formulate GNNs as follows:

\begin{equation}
\boldsymbol{\mathsf{h}}_\mathsf{v}^{(k+1)} = \textsc{Aggregation}\left(\left\{\boldsymbol{\mathsf{h}}^{(k)}_\mathsf{u}\right\}_{\mathsf{u} \in \mathcal{N}_\mathsf{v}\cup\left\{\mathsf{v}\right\}}\right) 
\end{equation}

\noindent where $\boldsymbol{\mathsf{h}}_\mathsf{v}^{(k)}$ is the vector representation of node $\mathsf{v}$ at the $k$-th layer; 
and $\mathcal{N}_\mathsf{v}$ is the set of neighbours of node $\mathsf{v}$.

There have been many designs for the \textsc{Aggregation} functions. The widely-used one is introduced in Graph Convolutional Networks (GCNs) \citep{kipf2017semi} as:

\begin{equation}
\boldsymbol{\mathsf{h}}_{\mathsf{v}}^{(k+1)} = \mathsf{g}\left(\sum_{\mathsf{u} \in \mathcal{N}_\mathsf{v}\cup\left\{\mathsf{v}\right\}}a_{\mathsf{v},\mathsf{u}}\boldsymbol{W}^{(k)}\boldsymbol{\mathsf{h}}_{\mathsf{u}}^{(k)}\right) , \forall \mathsf{v} \in \mathcal{V} 
\label{equa:gcn}
\end{equation}

\noindent where  
$\mathsf{g}$ is a nonlinear activation function such as $\mathsf{ReLU}$; $\boldsymbol{W}^{(k)}$ is a weight matrix at the $k$-th layer; and 
$a_{\mathsf{v},\mathsf{u}}$ is an edge constant between nodes $\mathsf{v}$ and $\mathsf{u}$ in the re-normalized adjacency matrix $\tilde{\textbf{D}}^{-\frac{1}{2}}\tilde{\textbf{A}}\tilde{\textbf{D}}^{-\frac{1}{2}}$, wherein $\tilde{\textbf{A}} = \textbf{A} + \textbf{I}$ where $\textbf{A}$ is the adjacency matrix, $\textbf{I}$ is the identity matrix, and  $\tilde{\textbf{D}}$ is the diagonal node degree matrix of $\tilde{\textbf{A}}$.

It is worth mentioning that several KG embedding approaches have been proposed to adapt GNNs for knowledge graph link prediction \citep{schlichtkrull2017modeling,shang2019end,vashishth2020compositionbased}. For example, R-GCN \citep{schlichtkrull2017modeling} modifies the basic form of GCNs  to introduce a specific encoder to update entity embeddings:

{
\begin{equation}
\boldsymbol{\mathsf{h}}_{\mathsf{e}}^{(k+1)} = \mathsf{g}\left(\sum_{r \in \mathcal{R}}\sum_{\mathsf{e'} \in \mathcal{N}_\mathsf{e}^r}\frac{1}{|\mathcal{N}_\mathsf{e}^r|}\boldsymbol{W}_r^{(k)}\boldsymbol{\mathsf{h}}_{\mathsf{e'}}^{(k)} + \boldsymbol{W}^{(k)}\boldsymbol{\mathsf{h}}_{\mathsf{e}}^{(k)}\right) 
\label{eqn:grcn}
\end{equation}
}

\noindent where $\mathcal{R}$ is a set of relations in the KG; $\mathcal{N}_\mathsf{e}^r = \left\{e' | ( e, r, e' ) \in \mathcal{T} \cup ( e', r, e ) \in \mathcal{T} \right\}$ 
denotes the set of entity neighbors of entity $\mathsf{e}$ via relation edge $r$, wherein $\mathcal{T}$ denotes the set of knowledge graph triples; and $\boldsymbol{W}_r^{(k)}$ is a weight transformation matrix associated with $r$ at the $k$-th layer. Then R-GCN uses DistMult \citep{Yang2015} as its decoder module to compute the score of \textit{(h, r, t)} as:

\begin{equation}
f\left(h, r, t\right) = \left<\boldsymbol{\mathsf{h}}_h^{(K)},\boldsymbol{v}_r,\boldsymbol{\mathsf{h}}_t^{(K)}\right> 
\label{equa:DistMult}
\end{equation}

\noindent where $\boldsymbol{\mathsf{h}}_h^{(K)}$ and $\boldsymbol{\mathsf{h}}_t^{(K)}$ are output vectors taken from the last layer of the encoder module; $\boldsymbol{v}_r$ denotes the embedding of relation $r$; and $\left<\right>$ denotes a multiple-linear dot product $\left<\boldsymbol{\mathsf{a}}, \boldsymbol{\mathsf{b}}, \boldsymbol{\mathsf{c}}\right> = \sum_i^n \boldsymbol{\mathsf{a}}_i  \times \boldsymbol{\mathsf{b}}_i \times  \boldsymbol{\mathsf{c}}_i$.

CompGCN \citep{vashishth2020compositionbased} also customizes GCNs to consider composition operations between entities and relations in the encoder module as follows:

\begin{align}
\boldsymbol{\mathsf{h}}_{\mathsf{e}}^{(k+1)} &= \mathsf{g}\left(\sum_{(\mathsf{e'}, r) \in \mathcal{N}_\mathsf{e}}
\boldsymbol{W}_{\mathsf{type}(r)}^{(k)}\phi\left(\boldsymbol{\mathsf{h}}_{\mathsf{e'}}^{(k)}, \boldsymbol{\mathsf{h}}_r^{(k)}\right)\right) \label{eqn:compgcn}  \\
\boldsymbol{\mathsf{h}}_r^{(k+1)} &= \boldsymbol{W}^{(k)}\boldsymbol{\mathsf{h}}_r^{(k)}  \label{eqn:compgcn2}
\end{align}

\noindent where $\mathcal{N}_\mathsf{e} = \left\{(e', r) | ( e, r, e' ) \in \mathcal{T} \cup ( e', r, e ) \in \mathcal{T} \right\}$ is the neighboring entity-relation pair set of entity $e$; and 
$\boldsymbol{W}_{\mathsf{type}(r)}^{(k)}$ denotes relation-type specific weight matrix. 
CompGCN explores the composition functions ($\phi$) inspired from TransE \citep{NIPS2013_5071}, DistMult, and HolE \citep{nickel2016holographic}.
Then CompGCN  uses ConvE \citep{Dettmers2017} as the decoder module.

The existing GNN-based KG embedding models, e.g. R-GCN  and CompGCN, mainly capture the graph structure surrounding entities. That is, as shown in Equations \ref{eqn:grcn}, \ref{eqn:compgcn} and \ref{eqn:compgcn2}, a relation's representation is \textit{not} directly used to update another relation's representation and is only used to update entity embeddings, while entity embeddings are \textit{not} used to update relation representations. Thus, these models might miss covering potentially useful relation structure information that is illustrated by the example (born\_in, Sydney, city\_in) in Section \ref{sec:intro}. 

\section{Our model WGE}\label{sec:wgemodel}

A knowledge graph $G = \left\{\mathcal{V}, \mathcal{R}, \mathcal{T}\right\}$ can be represented as a collection of factual valid triples \textit{(head entity, relation, tail entity)} denoted as $(h, r, t) \in \mathcal{T}$ with $h,t \in {\mathcal{V}}$ and $r \in \mathcal{R}$, wherein $\mathcal{V}$, $\mathcal{R}$ and $\mathcal{T}$ denote  the sets of entities, relations and triples, respectively.

\begin{figure*}[!t]
	\centering
	\includegraphics[width=0.99\linewidth]{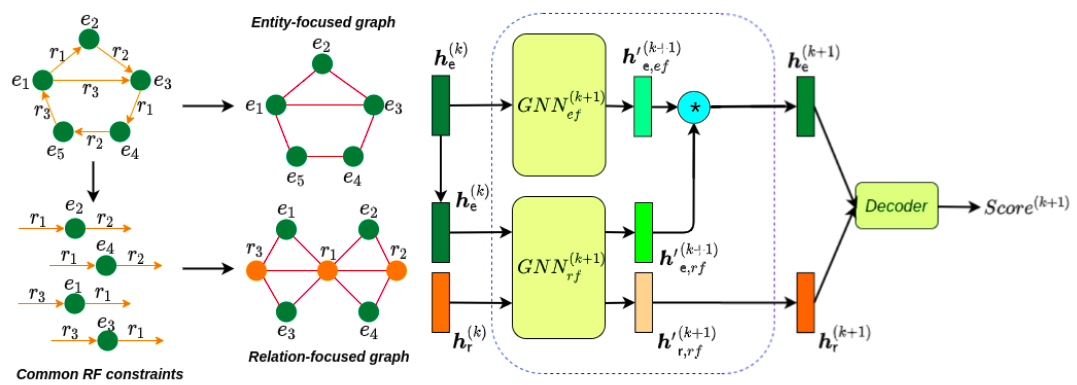}
    \caption{An illustration of our proposed WGE. Here, $\boldsymbol{h}^{(k)}_\mathsf{e}$ and $\boldsymbol{h}^{(k)}_r$ the vector representations of the entity $\mathsf{e}$ and the relation $r$ at $k$-th layer of the encoder module, are computed following Equation \ref{equal:final}.}
    \label{fig:graph_complex}
 \end{figure*}

To better capture the graph structure, as illustrated in Figure \ref{fig:graph_complex}, we introduce WGE as follows: (i) WGE transforms a given KG into two views: a single undirected entity-focused graph and a single undirected relation-focused graph. (ii) WGE introduces a new encoder architecture to update vector representations of entities and relations based on these two single graphs. (iii) WGE utilizes a weighted score function as the decoder module to compute the triple scores.

\subsection{Two-view construction}
\label{subsec:graphconstruction}

\subsubsection{Entity-focused view\ } WGE aims to obtain the entity neighborhood information. Thus, given a KG $G$, WGE constructs a single undirected graph $\mathcal{G}_{ef}$ viewing entities as individual nodes.
Here, $\mathcal{G}_{ef} = \left\{\mathcal{V}_{ef}, \mathcal{E}_{ef}\right\}$, wherein $\mathcal{V}_{ef}$ is the set of nodes and $\mathcal{E}_{ef}$ is the set of edges. 
The number of nodes in  $\mathcal{G}_{ef}$ is equal to the number of entities in $G$, i.e., $|\mathcal{V}_{ef}| = |\mathcal{V}|$.
In particular, for each triple $(h, r, t)$ in $G$, entities $h$ and $t$ become individual nodes in $\mathcal{G}_{ef}$ with an edge between them, as illustrated in Figure \ref{fig:graph_complex}. Here, 
$\mathcal{G}_{ef}$ is associated with an adjacency matrix $\boldsymbol{A}_{ef}$:

\begin{equation}
    \boldsymbol{A}_{ef}(\mathsf{v},\mathsf{u}) = \left\{ 
        \begin{array}{l}
            1 \ \ \ \text{if there is an edge between entity nodes $\mathsf{v}$ and $\mathsf{u}$} \\
            0 \ \ \ \text{otherwise }
        \end{array} \right. 
\end{equation}

\subsubsection{Relation-focused view\ } WGE also aims to attain the potential dependence between two neighborhood relations (e.g. ``child\_of'' and ``spouse'') to enhance learning representations.
To do that, from $G$, our WGE extracts relation-focused (RF) constraints in the form of (\textit{subjective relation, predicate entity, objective relation}), denoted as $(r_s, e_p, r_o)$, wherein $e_p$ is the tail entity for the relation $r_s$ and also the head entity for the relation $r_o$, e.g. (born\_in, Sydney, city\_in).
Here, WGE keeps a certain fraction $\beta$ of common RF constraints based on ranking how often two relations $r_s$ and $r_o$ co-appear in all extracted RF ones.
Then, WGE transforms those common obtained RF constraints into a single undirected relation-focused graph $\mathcal{G}_{rf} = \left\{\mathcal{V}_{rf}, \mathcal{E}_{rf}\right\}$ that views both entities and relations as individual nodes, wherein $\mathcal{V}_{rf}$ is the set of entity and relation nodes, $\mathcal{E}_{rf}$ is the set of edges. For example, as shown in Figure \ref{fig:graph_complex}, given an RF constraint $(r_1, e_2, r_2)$, WGE considers $r_1$, $e_2$, and $r_2$ as individual nodes in $\mathcal{G}_{rf}$ with edges among them.
$\mathcal{G}_{rf}$ is associated with an adjacency matrix $\boldsymbol{A}_{rf}$:

\begin{equation}
    \boldsymbol{A}_{rf}(\mathsf{v},\mathsf{u}) = \left\{ 
        \begin{array}{l}
            1 \ \ \ \text{if there is an edge between nodes $\mathsf{v}$ and $\mathsf{u}$} \\
            0 \ \ \ \text{otherwise }
        \end{array} \right. 
\end{equation}

\subsection{Encoder module}

Given a single graph $\mathcal{G} = \left(\mathcal{V}, \mathcal{E}\right)$, we might adopt vanilla GNNs or GCNs directly on $\mathcal{G}$ and its adjacency matrix $\boldsymbol{A}$ to learn node embeddings.
Recently, QGNN---Quaternion Graph Neural Network \citep{Nguyen2020QGNN}---has been proposed to learn node embeddings in the quaternion space as follows:

\begin{equation}
\boldsymbol{h}_{\mathsf{v}}^{(k+1),Q} = \mathsf{g}\left(\sum_{\mathsf{u} \in \mathcal{N}_\mathsf{v}\cup\left\{\mathsf{v}\right\}}\mathsf{a}_{\mathsf{v},\mathsf{u}}\boldsymbol{W}^{(k),Q}\otimes\boldsymbol{h}_{\mathsf{u}}^{(k),Q}\right) 
\label{equa:QGNN}
\end{equation}

\noindent where the superscript $^Q$ denotes the quaternion space; $k$ is the layer index; $\mathcal{N}_\mathsf{v}$ is the set of neighbors of node $\mathsf{v}$; $\boldsymbol{W}^{(k),Q}$ is a quaternion weight matrix;  
$\otimes$ denotes the Hamilton product; and $\mathsf{g}$ is a nonlinear activation function such as $\mathsf{tanh}$; $\boldsymbol{h}_\mathsf{u}^{(0),Q} \in \mathbb{H}^n$ is an input embedding vector for node $\mathsf{u}$, which is randomly initialized and updated during training; and  $\mathsf{a}_{\mathsf{v},\mathsf{u}}$ is an edge constant between nodes $\mathsf{v}$ and $\mathsf{u}$ in the Laplacian re-normalized adjacency matrix $\tilde{\textbf{D}}^{-\frac{1}{2}}\tilde{\textbf{A}}\tilde{\textbf{D}}^{-\frac{1}{2}}$ with
$\tilde{\textbf{A}} = \boldsymbol{A} + \textbf{I}$,
\label{equa:adjmatrix}
where $\boldsymbol{A}$ is the adjacency matrix, $\textbf{I}$ is the identity matrix, and  $\tilde{\textbf{D}}$ is the diagonal node degree matrix of $\tilde{\textbf{A}}$. See quaternion algebra  background in the Appendix.
QGNN has demonstrated its superior performances for downstream tasks such as graph classification and node classification.

Our WGE thus proposes a new  encoder architecture to learn entity and relation vector representations based on two different QGNNs, as illustrated in Figure \ref{fig:graph_complex}.
This new encoder aims to capture both entity- and relation-focused graph structures to better update vector representations for entities and relations as follows:

{
\begin{equation}
\boldsymbol{h'}_{\mathsf{v},ef}^{(k+1),Q} = \mathsf{g}\left(\sum_{\mathsf{u} \in \mathcal{N}_\mathsf{v}\cup\left\{\mathsf{v}\right\}}\mathsf{a}_{\mathsf{v},\mathsf{u},ef}\boldsymbol{W}_{ef}^{(k),Q}\otimes\boldsymbol{h}_{\mathsf{u},ef}^{(k),Q}\right)
\label{equa:QGNNef}
\end{equation}
}

\noindent where the subscript $_{ef}$ denotes for QGNN on the entity-focused graph $\mathcal{G}_{ef}$, and we define $\boldsymbol{h}_{\mathsf{u},ef}^{(k),Q}$ as:

\begin{equation}
\boldsymbol{h}_{\mathsf{u},ef}^{(k),Q} = 
\boldsymbol{h'}_{\mathsf{u},ef}^{(k),Q} * \boldsymbol{h'}_{\mathsf{u},rf}^{(k),Q} 
\label{equa:equahuef}
\end{equation}

\noindent where $*$ denotes a quaternion element-wise product, and $\boldsymbol{h'}_{\mathsf{u},rf}^{(k),Q}$ is computed following the Equation \ref{equa:QGNNrf}:

{
\begin{equation}
\boldsymbol{h'}_{\mathsf{v},rf}^{(k+1),Q} = \mathsf{g}\left(\sum_{\mathsf{u} \in \mathcal{N}_\mathsf{v}\cup\left\{\mathsf{v}\right\}}\mathsf{a}_{\mathsf{v},\mathsf{u},rf}\boldsymbol{W}_{rf}^{(k),Q}\otimes\boldsymbol{h}_{\mathsf{u}}^{(k),Q}\right) 
\label{equa:QGNNrf}
\end{equation}
}

\noindent where the subscript $_{rf}$ denotes for QGNN on the relation-focused graph $\mathcal{G}_{rf}$.
We define $\boldsymbol{h}_{\mathsf{u}}^{(k),Q}$ as:

{
\begin{equation}
    \boldsymbol{h}_{\mathsf{u}}^{(k),Q} = \left\{ 
        \begin{array}{l}
            \boldsymbol{h}_{\mathsf{u},ef}^{(k),Q} \ \ \ \text{if $\mathsf{u}$ is an entity node, as in Equation \ref{equa:equahuef} }
            \\
            \\
            \boldsymbol{h'}_{\mathsf{u},rf}^{(k),Q} \ \ \ \text{if $\mathsf{u}$ is a relation node, following Equation \ref{equa:QGNNrf} }
        \end{array} \right. 
        \label{equal:final}
\end{equation}
}

WGE uses $\boldsymbol{h}_{\mathsf{e}}^{(k),Q}$ and $\boldsymbol{h}_{\mathsf{r}}^{(k),Q}$ as computed following Equation \ref{equal:final} as the vector representations for entity $\mathsf{e}$ and relation $\mathsf{r}$ at the $k$-th layer of our encoder module, respectively. These vectors will be used as input for the decoder module.

Note that our encoder module is not merely using such a GNN but proposes a new manner where the two GNNs interact with each other to jointly learn entity and relation representations from two graphs. This interaction is crucial and novel and is directly responsible for the good performance of our model, showing that two-view modeling helps produce better scores than single-view modeling (See our ablation study in Section \ref{ssec:ablation}).

\subsection{Decoder module}

As the encoder module learns quaternion entity and relation embeddings, WGE employs the quaternion KG embedding model QuatE \citep{zhang2019quaternion} across all hidden layers of the encoder module to return a final score $f(h, r, t)$ for each triple $(h, r, t)$ as:

\begin{align}
    f_k(h, r, t) &= \left(\boldsymbol{h}_h^{(k),Q} \otimes \boldsymbol{h}_r^{\triangleleft,(k),Q}\right)\bullet \boldsymbol{h}_t^{(k),Q}   \\
    f(h, r, t) &= \sum_k \alpha_k f_k(h, r, t) 
\end{align}

\noindent where $\alpha_k \in [0, 1]$ is a fixed important weight of the $k$-th layer with $\sum_k \alpha_k = 1$; $\boldsymbol{h}_h^{(k),Q}$, $\boldsymbol{h}_r^{(k), Q}$, and $\boldsymbol{h}_t^{(k), Q}$ are quaternion vectors taken from the $k$-th layer of the encoder; $\otimes$, $^\triangleleft$ and $\bullet$ denote the Hamilton product, the normalized quaternion and  the quaternion-inner product, respectively.

\subsection{Objective function}
We train WGE by using Adam \citep{kingma2014adam} to optimize a weighted loss function as:

\begin{align}
&\mathcal{L} = -\sum_{\substack{(h, r, t) \in \{\mathcal{T} \cup \mathcal{T}'\}}} \sum_k \alpha_k\Big(l_{(h, r, t)}\log\big(p_{k}(h, r, t)\big) \nonumber  \\
& \ \ \ \ \ \ \ \ \ \ \  \ \ \ \  \ \ \ \ \ + \big(1 - l_{(h, r, t)}\big)\log\big(1 - p_{k}(h, r, t)\big)\Big)  
\label{equal:losscrossentropy}
\\
&\text{in which, } l_{(h, r, t)} = \left\{ 
  \begin{array}{l}
1 \ \ \ \text{for } (h, r, t)\in\mathcal{T} \\
0 \ \ \ \text{for } (h, r, t)\in\mathcal{T}' 
  \end{array} \right. \nonumber \\
& \text{and } p_{k}(h, r, t) = \mathsf{sigmoid}\big(f_k(h, r, t)\big) \nonumber
\end{align}

\noindent here, $\mathcal{T}$ and $\mathcal{T}'$ are collections of valid and invalid triples, respectively. $\mathcal{T}'$ is collected by corrupting valid triples in $\mathcal{T}$.

\section{Experiments}\label{sec:Exp}

We evaluate our proposed WGE for the KG completion task, i.e., link prediction \citep{NIPS2013_5071}, which aims to predict a missing entity given a relation with another entity, e.g., predicting a head entity $h$ given $(?, r, t)$ or predicting a tail entity $t$ given $(h, r, ?)$. 
The results are calculated by ranking the scores produced by the score function $f$ on triples in the test set.

\subsection{Setup}

\subsubsection{Datasets\ } 
Recent works \citep{safavi2020codex,LiterallyWikidata} show that there are some quality issues with previous existing KG completion datasets. For example, a large percentage of relations in FB15K-237 \citep{toutanova-chen-2015-observed} could be covered by a trivial frequency rule \citep{safavi2020codex}. Hence, they introduce six new KG completion benchmarks, consisting of CoDEx-S, CoDEx-M, CoDEx-L,\footnote{\url{https://github.com/tsafavi/codex} \citep{safavi2020codex}} LitWD1K, LitWD19K and LitWD48K.\footnote{\url{https://github.com/GenetAsefa/LiterallyWikidata} \citep{LiterallyWikidata}} These datasets are more difficult and cover more diverse and interpretable content than the previous ones. We use the six new challenging datasets as well as the FB15K-237 dataset to compare different models.
The statistics of these datasets are presented in Table \ref{tab:datasets}.

\begin{table}[!t]
\centering
\caption{Statistics of the experimental  datasets.}
\setlength{\tabcolsep}{0.4em}
\begin{tabular}{l|l|l|lll}
\hline
\multirow{2}{*}{\bf Dataset} & \multirow{2}{*}{\ \ \ $|\mathcal{E}|$} & \multirow{2}{*}{$|\mathcal{R}|$} & \multicolumn{3}{c}{$\#$Triples} \\
\cline{4-6}
& & &   Train & Valid & Test    \\
\hline
CoDEx-S & 2,034 & 42 & 32,888 & 1827 & 1828 \\
CoDEx-M & 17,050 & 51 & 185,584 & 10,310 & 10,311 \\
CoDEx-L & 77,951 & 69 & 551,193 & 30,622 & 30,622 \\
LitWD1K & 1,533 & 47 & 26,115 & 1,451 & 1,451 \\
LitWD19K & 18,986 & 182 & 260,039 & 14,447 & 14,447 \\
LitWD48K & 47,998 & 257 & 303,117 & 16,838 & 16,838 \\
FB15K-237 & 14,541 & 237 & 272,115 & 17,535 & 20,466 \\
\hline
\end{tabular}
\label{tab:datasets}
\end{table}

\subsubsection{Evaluation protocol\ } 
Following the standard protocol \citep{NIPS2013_5071}, to generate corrupted triples for each test triple $(h, r, t)$, we replace either $h$ or $t$ by each of all other entities in turn. We also apply the ``{Filtered}'' setting protocol \citep{NIPS2013_5071} to filter out before ranking any corrupted triples that appear in the KG. We then rank the valid test triple as well as the corrupted triples in descending order of their triple scores. We report standard evaluation metrics: mean reciprocal rank (MRR) and Hits@10 (i.e. the proportion of test triples for which the target entity is ranked in the top 10 predictions). Here, a higher MRR/Hits@10 score reflects a better prediction result.

\subsubsection{Our model's training protocol\ } 
We implement our model using  Pytorch \citep{NEURIPS2019_9015}.
We apply the standard Glorot initialization \citep{glorot2010understanding} for parameter initialization. 
We employ $\mathsf{tanh}$ for the nonlinear activation function $\mathsf{g}$.
We use the Adam optimizer \citep{kingma2014adam} to train our WGE model up to 3000 epochs on all datasets. We use a grid search to choose the number $K$ of hidden layers $\in \{1, 2, 3\}$, the Adam initial learning rate  $\in \left\{1e^{-4}, 5e^{-4}, 1e^{-3}, 5e^{-3}\right\}$, the batch size $\in \left\{1024, 2048, 4096\right\}$, and the input dimension and hidden sizes  of the QGNN hidden layers $\in \left\{32, 64, 128, 256, 512, 1024 \right\}$. For the decoder module, we perform a grid search to select its mixture weight value $\alpha_0 \in \{0.3, 0.6, 0.9\}$, and fix the mixture weight values for the $K$ layers at  $\alpha_k=\dfrac{1-\alpha_0}{K}$. 
For the percentage $\beta$ of kept RF constraints, we  grid-search $\beta \in \{0.1, 0.2, ..., 0.9\}$ for the CoDEx-S dataset, and the best value is $0.2$; then we use $\beta=0.2$ \underline{for all remaining datasets}. We evaluate the MRR after every 10 training epochs on the validation set to select the best model checkpoint, and then apply the selected one to the test set.

\subsubsection{Baselines' training protocol\ }   
For strong baseline models, we apply the same evaluation protocol. The training protocol is the same w.r.t. parameter initialization, the optimizer, the hidden layers, the initial learning rate values, the batch sizes and the number of training epochs as well as the best model checkpoint selection. We also use a model-specific configuration for each baseline. In particular, for {TransE} \citep{NIPS2013_5071}, {ConvE} \citep{Dettmers2017}, {TuckER} \citep{balazevic2019tucker} and {QuatE}, we use grid search to choose the embedding dimension in \{64, 128, 256, 512\}. For the QGNN-based KG embedding model {SimQGNN} \citep{Nguyen2020QGNN} that obtains state-of-the-art results on the CoDEx datasets, we successfully reproduce this model's reported results using its optimal hyper-parameters. For R-GCN and CompGCN, we use 2 GCN layers and vary the embedding size of the GCN layer from \{64, 128, 256, 512\}.  
For WGE variants in the Ablation study, we also set the same dimension value for both the embedding size and the hidden size, wherein we vary the dimension value in \{64, 128, 256, 512\}.

\begin{table*}[!t]
\centering
\caption{Experimental results on seven \textit{test} sets. Hits@10 (H@10) is reported in \%. 
The best scores are in {bold}, while the second best scores are in {underline}.
The results of TransE \citep{NIPS2013_5071}, ComplEx \citep{Trouillon2016}, ConvE \citep{Dettmers2017} and TuckER \citep{balazevic2019tucker} 
on three CoDEx test sets  are taken from \cite{safavi2020codex}.  The results of  R-GCN \citep{schlichtkrull2017modeling} 
and CompGCN \citep{vashishth2020compositionbased} and SimQGNN \citep{Nguyen2020QGNN} on three CoDEx  test sets are taken from \cite{Nguyen2020QGNN}.  The ComplEx results on three LitWD test sets   are taken from \cite{LiterallyWikidata}. The results of TransE, ComplEx, ConvE, R-GCN  and CompGCN on the FB15K-237 test set are taken from \cite{vashishth2020compositionbased}. The results of TuckER on FB15K-237 are taken from \cite{balazevic2019tucker}.
All results are reported using the same setup.
}
\resizebox{12cm}{!}{
\begin{tabular}{l|cc|cc|cc|cc|cc|cc|cc}
\hline
\multirow{2}{*}{\bf Method} & \multicolumn{2}{c|}{\bf CoDEx-S} & \multicolumn{2}{c|}{\bf CoDEx-M} & \multicolumn{2}{c|}{\bf CoDEx-L} & \multicolumn{2}{c|}{\bf LitWD1K} & \multicolumn{2}{c|}{\bf LitWD19K} & \multicolumn{2}{c|}{\bf LitWD48K} &
\multicolumn{2}{c}{\bf FB15K-237} \\
\cline{2-15} 
& MRR   & H@10  & MRR   & H@10  & MRR   & H@10 & MRR   & H@10  & MRR   & H@10  & MRR   & H@10 & MRR   & H@10 \\
\hline
TransE  & 0.354 & 63.4 & 0.303 & 45.4 & 0.187 & 31.7 & 0.313 & 51.3 & 0.172 & 26.4 & 0.269 & 41.3 & 0.294 & 46.5 \\
ComplEx & \bf 0.465 & {64.6} & \underline{0.337} & {47.6} & 0.294 & 40.0 & 0.413 & 67.3 & 0.181 & 29.6 & 0.277 & 42.8 & 0.247 & 33.9\\
ConvE   & {0.444} & 63.5 & 0.318 & 46.4 & 0.303 & 42.0 & 0.477 & 71.4 & 0.310 & 45.1 & 0.372 & 54.0 & 0.325 & 50.1 \\
TuckER  & {0.444} & 63.8 & {0.328} & 45.8 & {0.309} & {43.0} & 0.498 & 74.4 & 0.311 & 46.3 & 0.391 & \underline{58.7} & \textbf{0.358} & \textbf{54.4}\\

\hline
R-GCN   & 0.275 & 53.3 & 0.124 & 24.1 & 0.073 & 14.2 & 0.244 & 46.2 & 0.211 & 34.1 & 0.238 & 44.2 & 0.248 & 41.7 \\
CompGCN & 0.395 & 62.1 & 0.312 & 45.7 & 0.304 & 42.8 & 0.323 & 52.8 & 0.319 & 47.4 & 0.379 & 58.4 & \underline{0.355} & {53.5} \\

SimQGNN  & 0.435 & \underline{65.2} & 0.323 & {47.7} & {0.310} & {43.7} & \underline{0.518} & \underline{75.1} & 0.308 & 46.9 & 0.350 & 57.6 & 0.339 & 51.8 \\
\hline

QuatE   & 0.449 & 64.4 & 0.323 & \underline{48.0} & \underline{0.312} & \underline{44.3} & 0.514 & 73.1 & \underline{0.341} & \underline{49.3} & \underline{0.392} & {58.6} & 0.342 & 52.9  \\
\hline
\textbf{WGE}  & \underline{0.452} & \bf 66.4 & \bf 0.338 & \bf 48.5 & \bf 0.320 & \bf 44.5 & \bf 0.527 & \bf 76.2 & \bf 0.345 & \bf  49.9 & \bf 0.401 & \bf 59.5 & {0.348} & \underline{53.6} \\

\hline
\end{tabular}
}

\label{tab:resultsCoDEx}
\end{table*}

\begin{figure*}[!t]
	\centering
	\subfigure[Effects of the percentage $\beta$.]{
    \includegraphics[width=0.475\linewidth]{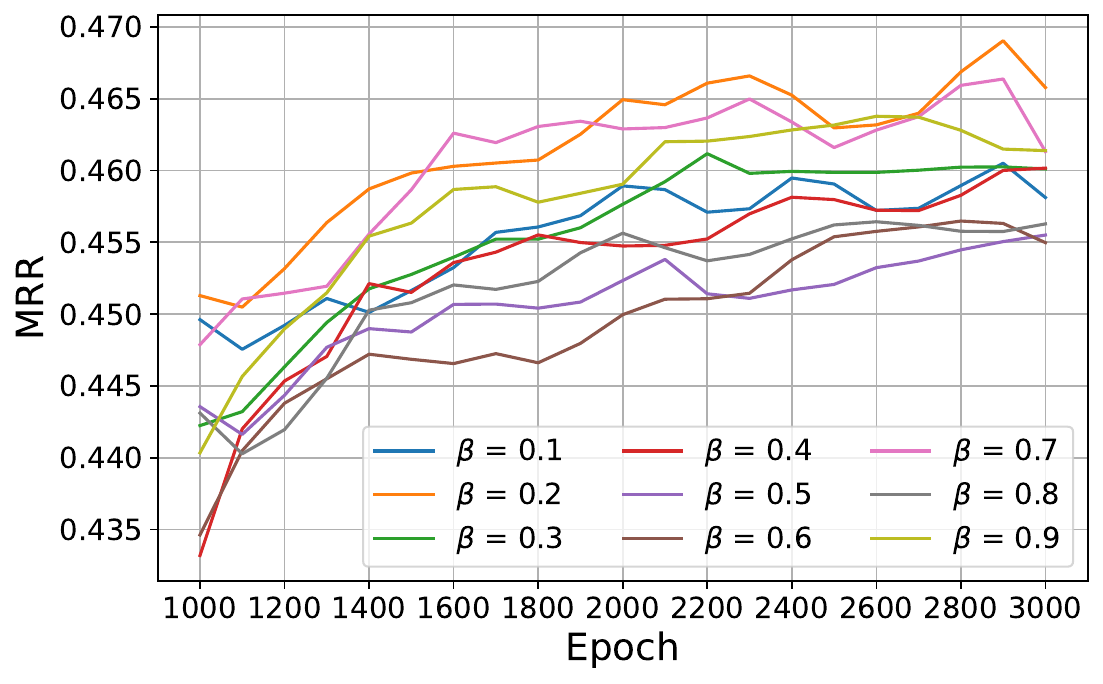}
    \label{subfig:beta}
    }
    \subfigure[Effects of the embedding sizes.]{
    \includegraphics[width=0.475\linewidth]{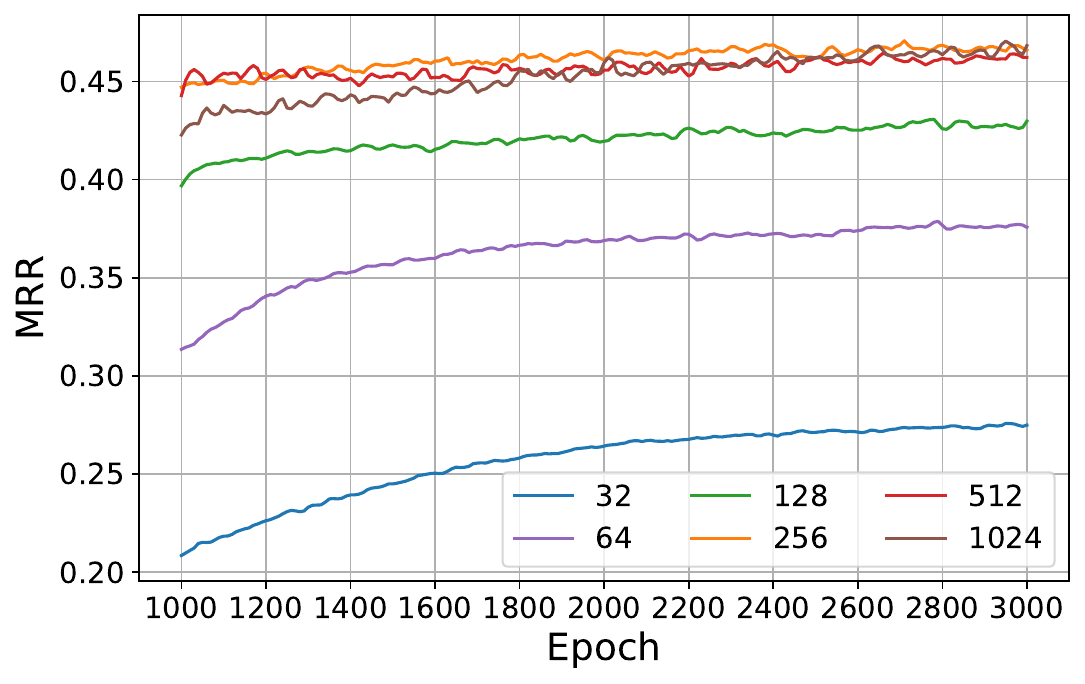}
    \label{subfig:dimsize}
    }
    \caption{Effects of hyper-parameters on the CoDEx-S validation set.}
\end{figure*}

\subsection{Main results} 
Table \ref{tab:resultsCoDEx} shows our results obtained for WGE and other strong baselines on seven experimental datasets. 
In general, our WGE obtains the highest MRR and Hits@10 scores on all three CoDEx and three LitDW challenge datasets (except the second highest MRR on CoDEx-S); and on FB15K-237, WGE obtains the third highest MRR and the second highest Hits@10. 
In particular, WGE gains substantial improvements compared to both R-GCN and CompGCN on all three CoDEx and three LitDW challenge datasets. Compared to the QGNN-based model SimQGNN, our WGE obtains 1.5\% and 0.02 absolute higher Hits@10 and MRR scores averaged over all seven datasets than SimQGNN, respectively. 
We also find that QuatE obtains competitive performance scores when carefully tuning its hyper-parameters (e.g. generally outperforming SimQGNN),\footnote{Note that the experimental setup is the same for both QuatE and WGE for a fair comparison as WGE uses QuatE for decoding. Zhang et al. \citep{zhang2019quaternion} reported  MRR at 0.348 and  Hits@10 at 55.0\% on FB15K-237 for QuatE. However, we could not reproduce those scores. }
however, it is still surpassed by WGE by about 1.1+\% and 0.01 on averaged Hits@10 and MRR, respectively.

\subsubsection{Hyper-parameter sensitivity\ } We present in Figures \ref{subfig:beta} and \ref{subfig:dimsize} the effects of essential hyper-parameters including the percentage $\beta$ of kept RF constraints and the embedding sizes on the CoDEx-S validation set.

\begin{itemize}
\item \textbf{Percentage $\beta$ of kept RF constraints}: As defined in Section \ref{subsec:graphconstruction}, the hyper-parameter $\beta$ aims to determine the number of common RF constraints to be kept in the relation-focused graph. 
We visualize the MRR scores according to the value of $\beta$ in $\{0.1, 0.2, ..., 0.9\}$ in Figure \ref{subfig:beta}.\footnote{Our training protocol  monitors the MRR score on the validation set to select the best model checkpoint.} 
We find that WGE performs best with $\beta = 0.2$. 
Recall that the hyper-parameter $\beta = 0.2$ is tuned on the CoDEx-S validation set only, and then used for all remaining datasets. Here, the hyper-parameter $\beta = 0.2$ already helps our WGE to outperform strong baselines, as shown in Table \ref{tab:resultsCoDEx}. Our scores obtained on the remaining datasets are likely better if $\beta$ is also tuned on those datasets. 
A limitation of our approach is that the mechanism of selecting kept RF constraints in the Relation-focused view is based on the observed co-occurrence frequency between entities and relations. This might not be optimal as some entity-relation pairs can have important interactions regardless of their small number of co-occurrences as the observed KG is incomplete (the actual number of co-occurrences could be larger). In future work, we would design a soft scoring mechanism that gives a score for each entity-relation pair and be able to adaptively prune the graph during training. 
    
\item \textbf{Embedding sizes}: 
Figure \ref{subfig:dimsize} illustrates the performance differences of WGE when varying the embedding size in $\{32, 64, 128, 256, 512, 1024\}$. 
Our WGE achieves the highest MRR when the embedding size is 256. 
We find that there are no substantial MRR gains when the size is larger than 256. We also observe similar findings for the remaining datasets.
\end{itemize}

\begin{figure*}[!t]
	\centering
	\subfigure[Tail prediction on CoDEx-S.]{
    \includegraphics[width=0.475\linewidth]{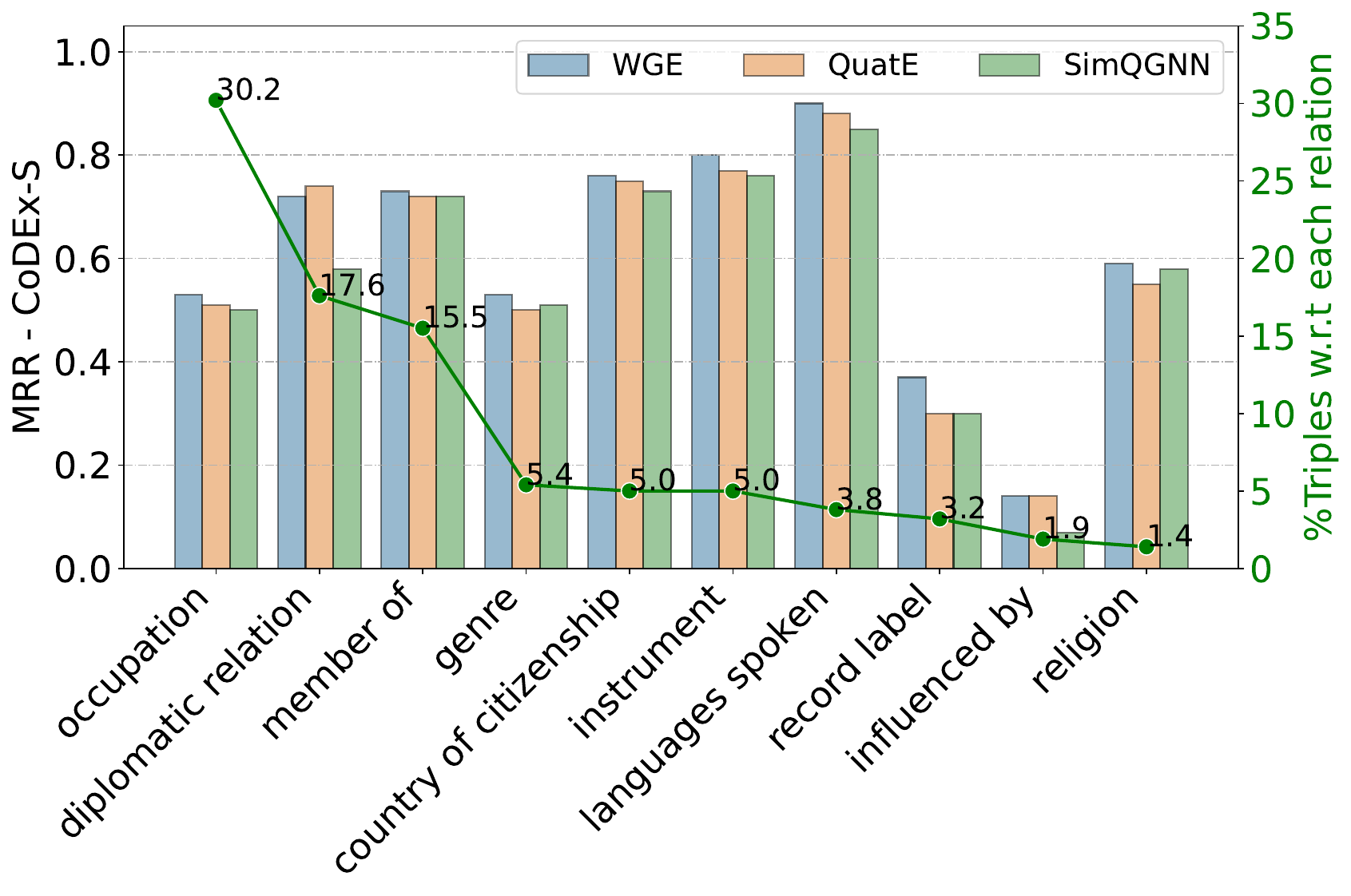}
    \label{subfig:codexstail}
    }
    \subfigure[Head prediction on CoDEx-S.]{
    \includegraphics[width=0.475\linewidth]{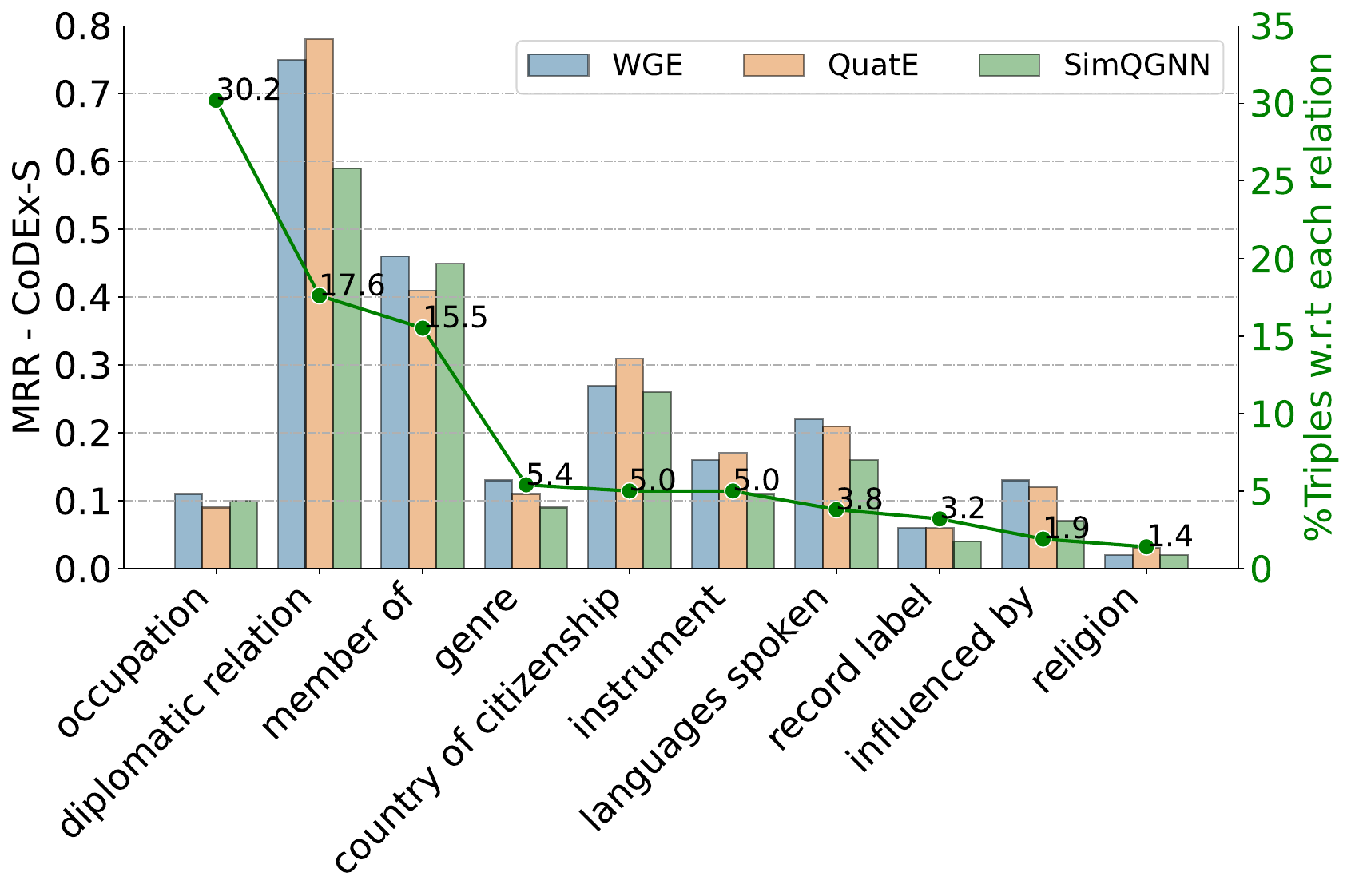}
    \label{subfig:codexshead}
    }

	\subfigure[Tail prediction on CoDEx-M.]{
    \includegraphics[width=0.475\linewidth]{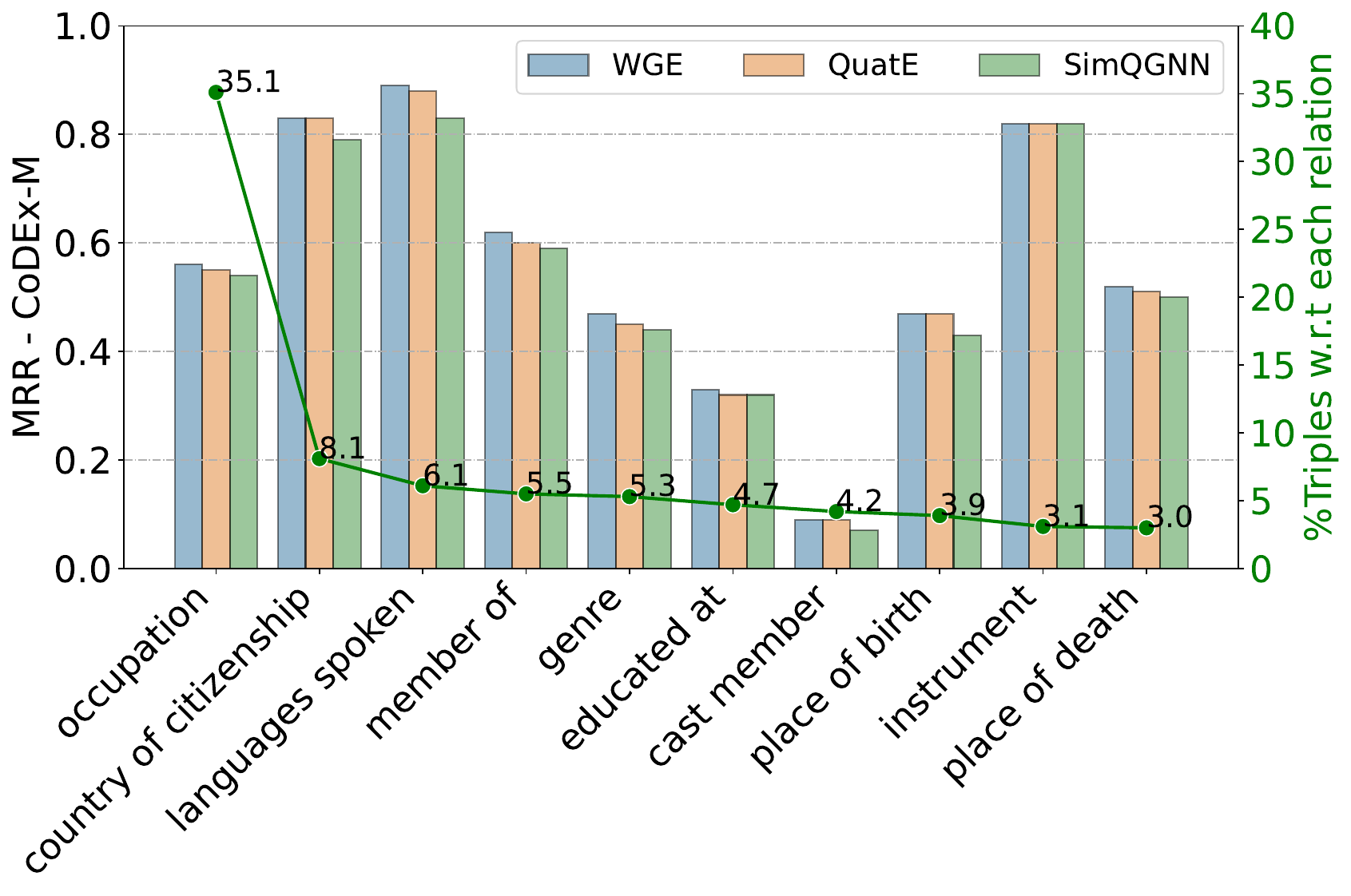}
    \label{subfig:codexmtail}
    }
    \subfigure[Head prediction on CoDEx-M.]{
    \includegraphics[width=0.475\linewidth]{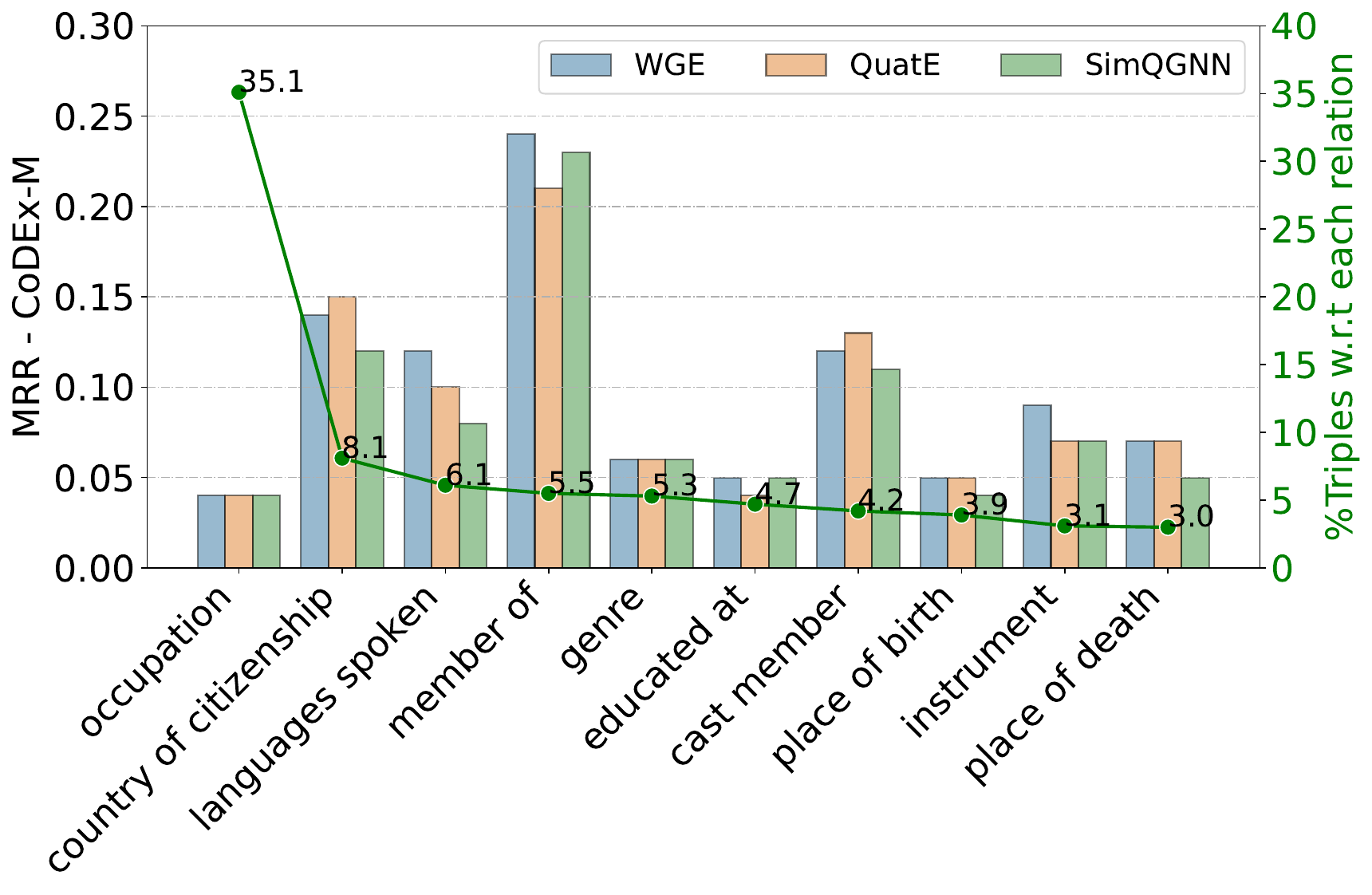}
    \label{subfig:codexmhead}
    }

    \caption{MRR on the CoDEx-S and CoDEx-M validation sets w.r.t each relation. The right y-axis is the percentage of triples corresponding to each relation.}
    \label{fig:codexlfig}
\end{figure*}

\begin{figure*}[!t]
	\centering

    \subfigure[Tail prediction on CoDEx-L.]{
    \includegraphics[width=0.475\linewidth]{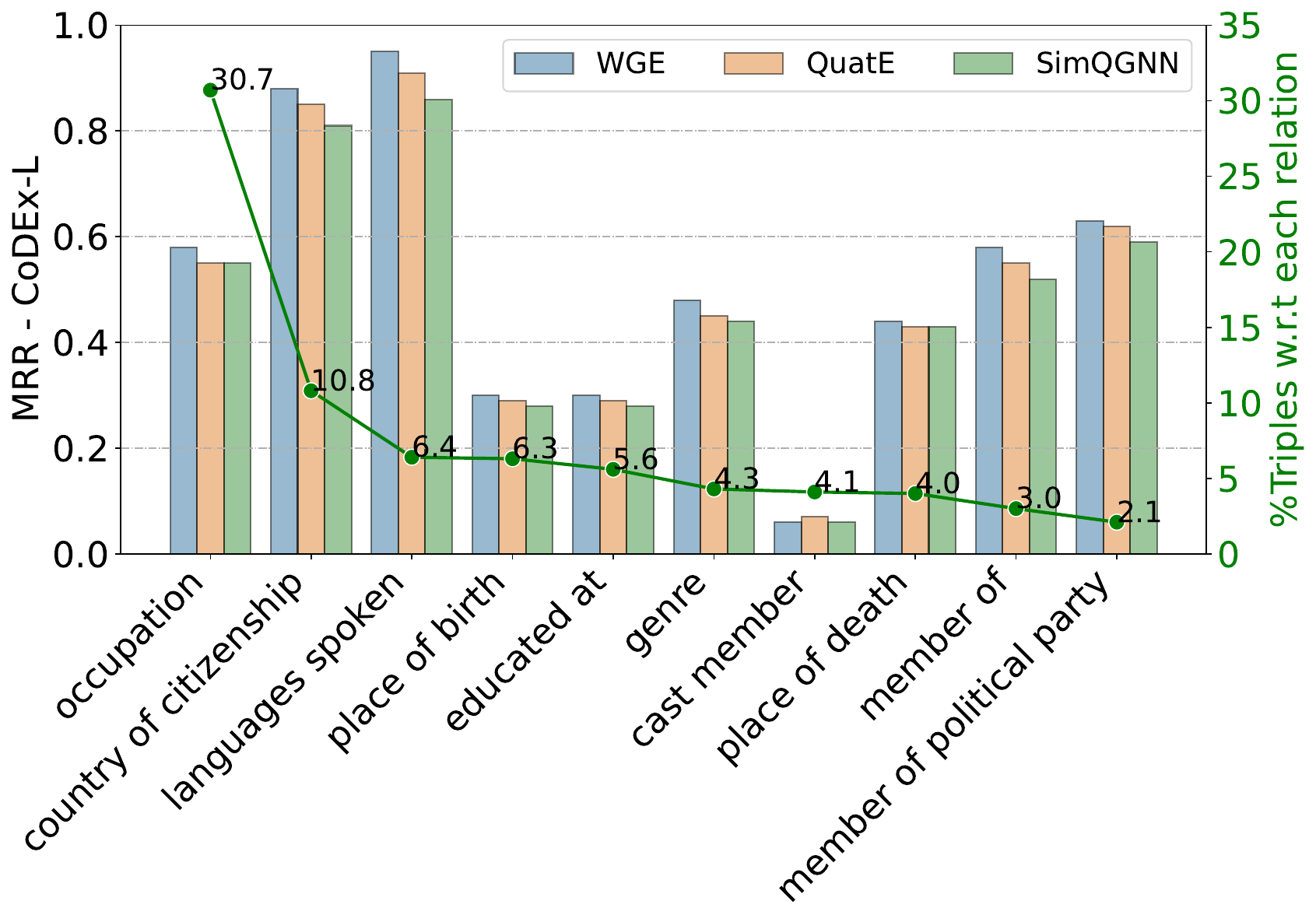}
    \label{subfig:codexltail}
    }
    \subfigure[Head prediction on CoDEx-L.]{
    \includegraphics[width=0.475\linewidth]{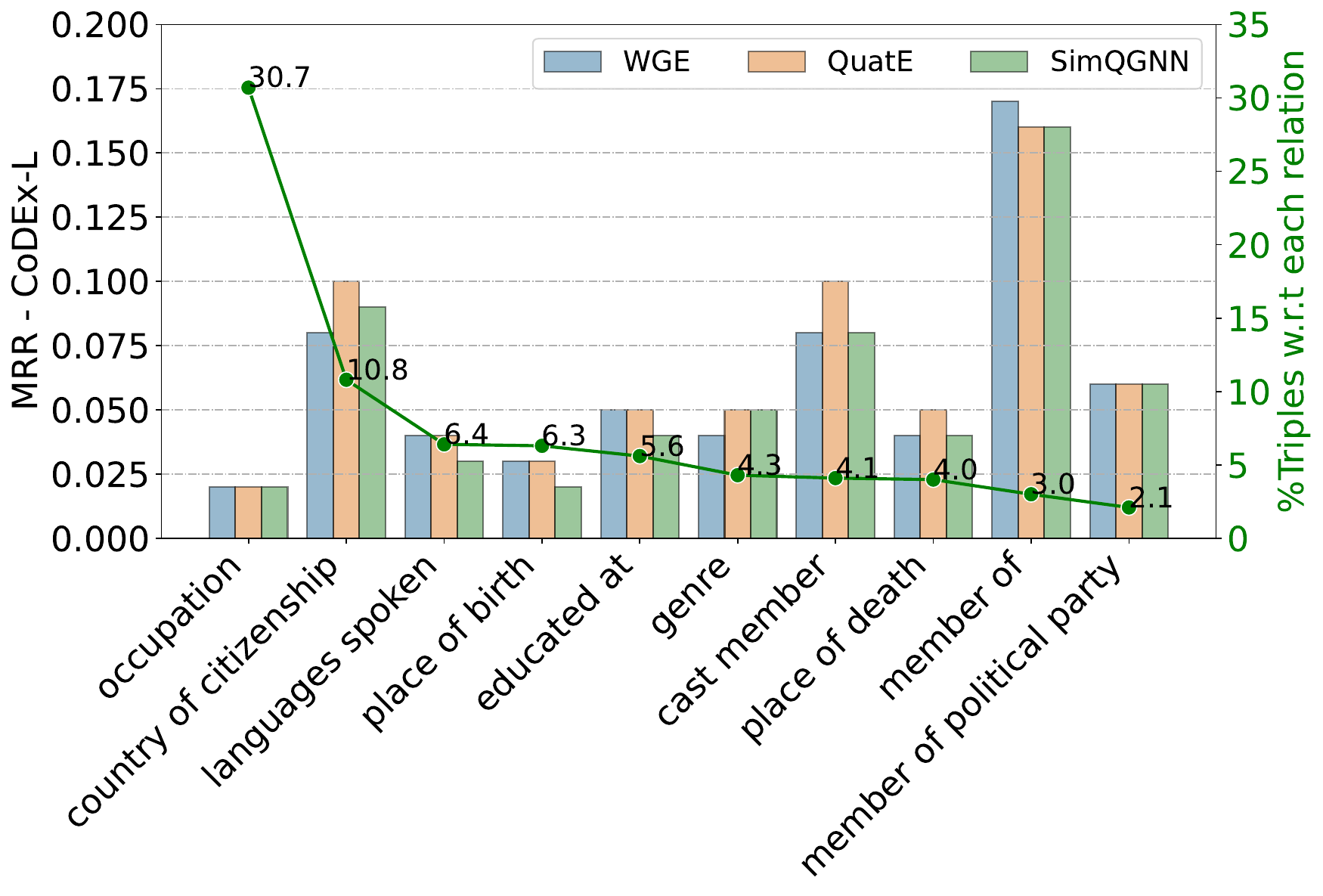}
    \label{subfig:codexlhead}
    }

    \caption{MRR on the CoDEx-L validation set w.r.t each relation. The right y-axis is the percentage of triples corresponding to each relation.}
    \label{fig:codexlfig2}
\end{figure*}

\subsubsection{Qualitative study\ } We report the performances of WGE, QuatE and SimQGNN over different relation types on the CoDEx validation sets in Figures  \ref{fig:codexlfig} and \ref{fig:codexlfig2}. 
For each dataset, we select the top 10 frequent relations and compare model performances over these 10 relations. 
We also separate the result into tail prediction (i.e., predicting the tail entity given $(h, r, ?)$) and head prediction (i.e., predicting the head entity given $(?, r, t)$). 
WGE generally works better than both QuatE and SimQGNN except for some special relation cases.
For example, QuatE achieves higher head prediction scores for the relation ``\textit{country of citizenship}'' than WGE as shown in Figures \ref{subfig:codexshead}, \ref{subfig:codexmhead} and \ref{subfig:codexlhead}.
A possible reason is that some useful RF constraints related to the relation ``\textit{country of citizenship}'' have been omitted from the relation-focused graph construction. 
Note that there is a substantial performance gap between the head prediction and the tail prediction, wherein predicting the tail entities is easier than predicting the head entities.
The reason might come from the fact that in the CoDEx datasets, each relation is associated with a small number of tail entities but with a large number of head entities. For example, head entity candidates for ``\textit{occupation}" relations can be any person nodes, while candidates for tail entities are limited by the number of job entities.

\subsection{Ablation analysis}\label{ssec:ablation}
Tables \ref{tab:ablationstudy} and \ref{tab:ablationstudy2}  present our ablation results on the validation sets for five variants of our proposed WGE, including:

\begin{itemize}

\item (1) \textbf{A variant without predicate entities}: This is a variant that only keeps relation nodes in the relation-focused view, i.e., \textit{without using the predicate entities} as nodes from the extracted RF constraints.

        \item (2) \textbf{A variant with GCN}: This is a variant that uses GCN in the encoder module instead of using QGNN.

    \item (3) \textbf{A variant with only entity-focused view}: This is a variant that uses only the entity-focused view. 

    \item (4) \textbf{A variant with only relation-focused view}: This is a variant that uses only the relation-focused view. 

    \item (5) \textbf{A variant with the Levi graph transformation}: This is a variant where a single Levi graph is used as the input of the encoder module. From the given KG, we investigate another strategy of constructing a single undirected graph, which can be considered as a direct extension of our entity-focused graph view with additional relation nodes, following the Levi graph transformation \citep{levi1942finite}. 

    
\end{itemize}

\begin{table*}[!t]
\centering
\caption{Ablation results on CoDEx {validation} sets for five variants of our WGE. 
(1) A variant where the relation-focused view uses only relation nodes, 
without using the predicate entities. 
(2) A variant utilizes GCN in the encoder module instead of using QGNN.
(3) A variant utilizes only the entity-focused view. 
(4) A variant utilizes only the relation-focused view.
(5) A variant uses the Levi graph transformation, i.e. the entity-focused graph view with addition relation nodes. 
}
\setlength{\tabcolsep}{0.2em}
\begin{tabular}{l|cc|cc|cc}
\hline
\multirow{2}{*}{\bf Method} & \multicolumn{2}{c|}{\bf CoDEx-S} & \multicolumn{2}{c|}{\bf CoDEx-M} & \multicolumn{2}{c}{\bf CoDEx-L} \\
\cline{2-7}     & MRR   & H@10  & MRR   & H@10  & MRR   & H@10   \\
\hline
\textbf{WGE}                                & \bf 0.469 & \bf 67.9 & \bf 0.339 & \bf 48.4 & \bf 0.320 & \bf 44.1  \\
\hdashline
\ \ \ \ (1) w/o predicate entities          & 0.448 & \underline{67.1} & 0.328 & \underline{47.1} & 0.312 & \underline{43.1} \\
\ \ \ \ (2) w/ GCN                          & 0.441 & 66.5 & 0.322 & 47.0 & 0.306 & {43.0}  \\
\ \ \ \ (3) w/ only entity-focused          & 0.452 & 66.9 & \underline{0.329} & 46.5 & \underline{0.314} & 43.0  \\
\ \ \ \ (4) w/ only relation-focused      & \underline{0.455} & 66.9 & 0.323 & 46.7 & 0.305 & 42.9 \\   
\ \ \ \ (5) w/ only Levi graph             & 0.447 & 63.5 & 0.320 & 45.7 & 0.288 & 41.1 \\
\hline
\end{tabular}
\label{tab:ablationstudy}
\end{table*}

\begin{table*}[!t]
\centering
\caption{Ablation results on LitWD and FB15K-237 {validation} sets for five variants of our WGE.
}
\resizebox{12cm}{!}{
\setlength{\tabcolsep}{0.2em}
\begin{tabular}{l|cc|cc|cc|cc}
\hline
\multirow{2}{*}{\bf Method} & \multicolumn{2}{c|}{\bf LitWD1K} & \multicolumn{2}{c|}{\bf LitWD19K} & \multicolumn{2}{c|}{\bf LitWD48K} &
\multicolumn{2}{c}{\bf FB15K-237} \\
\cline{2-9}     & MRR   & H@10  & MRR   & H@10  & MRR   & H@10 & MRR & H@10 \\
\hline
\textbf{WGE}                                & \bf 0.518 & \bf 75.5 & \bf 0.343 & \bf  49.5 & \bf 0.402 & \bf 59.3 & \bf 0.351 & \bf 53.6 \\
\hdashline
\ \ \ \ (1) w/o predicate entities          & 0.483 & 72.6 & 0.326 & 47.9 & 0.389 & 57.2 & 0.339 & 52.4 \\
\ \ \ \ (2) w/ GCN                          & 0.470 & 71.3 & 0.325 & 47.2 & 0.382 & 56.6 & 0.327 & 50.2 \\
\ \ \ \ (3) w/ only entity-focused          & 0.497 & 73.3 & 0.336 & 48.1 & \underline{0.397} & \underline{58.4} & \underline{0.341} & \underline{52.5} \\
\ \ \ \ (4) w/ only relation-focused      & \underline{0.498} & \underline{73.7} & \underline{0.338} & \underline{48.4} & 0.395 & \underline{58.4} & 0.340 & 52.3 \\   
\ \ \ \ (5) w/ only Levi graph             & 0.484 & 72.8 & 0.331 & 48.2 & 0.387 & 56.9 & 0.336 & 50.8 \\
\hline
\end{tabular}
}
\label{tab:ablationstudy2}
\end{table*}

We find that WGE outperforms all of its variants, thus showing that: from (1), the predicate entities can help to better infer the potential dependence between two neighborhood relations; from (2), GCNs are not as effective as QGNNs; and from (3), (4) and (5),  the modeling of two-view graphs of KGs helps produce better scores than single-view modeling of KGs, confirming the effectiveness of our two-view WGE approach. In addition, variant (5) obtains lower scores than variant (3), also showing that the Levi graph transformation is not as effective as the entity-focused graph transformation. 

\section{Conclusion}
\label{sec:conclusion}

In this paper, we have introduced WGE---an effective GNN-based KG embedding model---to enhance the entity neighborhood information with the potential dependence between two neighborhood relations. 
In particular, WGE constructs two views from the given KG, including a single undirected entity-focused graph and a single undirected relation-focused graph. 
Then WGE proposes a new encoder architecture to update entity and relation vector representations from these two graph views. 
After that, WGE employs a weighted score function to compute the triple scores for KG completion. 
Extensive experiments show that WGE outperforms other strong GNN-based baselines and KG embedding models on seven KG completion benchmark datasets. Our WGE implementation is publicly available at: \url{https://github.com/vinhsuhi/WGE}.

\section*{Acknowledgment} Most of this work was done while Vinh Tong was a research resident at VinAI Research, Vietnam.

\section*{Appendix}


The hyper-complex vector space has recently been considered on the Quaternion space \cite{hamilton1844ii} consisting of one real and three separate imaginary axes. 
It provides highly expressive computations through the Hamilton product compared to the Euclidean and complex vector spaces.
We provide key notations and operations related to the Quaternion space required for our later development. Additional details can further be found in \cite{parcollet2019survey}.

A quaternion $q \in \mathbb{H}$ is a hyper-complex number consisting of one real and three separate imaginary components \cite{hamilton1844ii} defined as:

\begin{equation}
q = q_r + q_i\boldsymbol{\mathsf{i}} + q_j\boldsymbol{\mathsf{j}} + q_k\boldsymbol{\mathsf{k}}  
\end{equation}

\noindent where $q_r, q_i, q_j, q_k \in \mathbb{R}$, and $\boldsymbol{\mathsf{i}}, \boldsymbol{\mathsf{j}}, \boldsymbol{\mathsf{k}}$ are imaginary units that $\boldsymbol{\mathsf{i}}^2 = \boldsymbol{\mathsf{j}}^2 = \boldsymbol{\mathsf{k}}^2 = \boldsymbol{\mathsf{i}}\boldsymbol{\mathsf{j}}\boldsymbol{\mathsf{k}} = -1$. 
The operations for the Quaternion algebra are defined as follows:

\textbf{Addition.} 
The addition of two quaternions $q$ and $p$ is defined as:

\begin{equation}
q + p = (q_r + p_r) + (q_i + p_i) \boldsymbol{\mathsf{i}} + (q_j + p_j)\boldsymbol{\mathsf{j}} + (q_k + p_k)\boldsymbol{\mathsf{k}} 
\end{equation}

\textbf{Norm.} The norm $\|q\|$ of a quaternion $q$ is computed as:

\begin{equation}
\|q\| = \sqrt{q_r^2 + q_i^2 + q_j^2 + q_k^2}  
\end{equation}
And the normalized or unit quaternion $q^\triangleleft$ is defined as:
$q^\triangleleft = \frac{q}{\|q\|}$

\textbf{Scalar multiplication.} 
The multiplication of a scalar $\lambda$ and $q$ is computed as follows:

\begin{equation}
\lambda q = \lambda q_r + \lambda q_i\boldsymbol{\mathsf{i}} + \lambda q_j\boldsymbol{\mathsf{j}} + \lambda q_k\boldsymbol{\mathsf{k}}  
\end{equation}

\textbf{Conjugate.} The conjugate $q^\ast$ of a quaternion $q$ is defined as:
\begin{equation}
q^\ast = q_r - q_i\boldsymbol{\mathsf{i}} - q_j\boldsymbol{\mathsf{j}} - q_k\boldsymbol{\mathsf{k}}  
\end{equation}

\textbf{Hamilton product.} The Hamilton product $\otimes$ (i.e., the quaternion multiplication) of two quaternions $q$ and $p$ is defined as:

\begin{align}
q \otimes p &=& (q_r p_r - q_i p_i - q_j p_j - q_k p_k) \nonumber \\
&+& (q_i p_r + q_r p_i - q_k p_j + q_j p_k)\boldsymbol{\mathsf{i}} \nonumber \\
&+& (q_j p_r + q_k p_i + q_r p_j - q_i p_k)\boldsymbol{\mathsf{j}} \nonumber \\
&+& (q_k p_r - q_j p_i + q_i p_j + q_r p_k)\boldsymbol{\mathsf{k}}   
\label{equa:halproduct}
\end{align}

\noindent We can express the Hamilton product of $q$ and $p$ in the following form:

\begin{equation}
q \otimes p = 
\begin{bmatrix}
1\\
\boldsymbol{\mathsf{i}}\\
\boldsymbol{\mathsf{j}}\\
\boldsymbol{\mathsf{k}}
\end{bmatrix}^\top
\begin{bmatrix}
q_r & -q_i & -q_j & -q_k\\
q_i & q_r & -q_k & q_j\\
q_j & q_k & q_r & -q_i\\
q_k & -q_j & q_i & q_r
\end{bmatrix}
\begin{bmatrix}
p_r\\
p_i\\
p_j\\
p_k
\end{bmatrix}   
\label{equa:halproduct_newform}
\end{equation}

\noindent  The Hamilton product of two quaternion vectors $\boldsymbol{q}$ and $\boldsymbol{p} \in \mathbb{H}^n$ is computed as:

\begin{align}
\boldsymbol{q} \otimes \boldsymbol{p} &=& (\boldsymbol{q}_r \circ \boldsymbol{p}_r - \boldsymbol{q}_i \circ \boldsymbol{p}_i - \boldsymbol{q}_j \circ \boldsymbol{p}_j - \boldsymbol{q}_k \circ \boldsymbol{p}_k)  \nonumber \\
&+& (\boldsymbol{q}_i \circ \boldsymbol{p}_r + \boldsymbol{q}_r \circ \boldsymbol{p}_i - \boldsymbol{q}_k \circ \boldsymbol{p}_j + \boldsymbol{q}_j \circ \boldsymbol{p}_k)\boldsymbol{\mathsf{i}} \nonumber \\
&+& (\boldsymbol{q}_j \circ \boldsymbol{p}_r + \boldsymbol{q}_k \circ \boldsymbol{p}_i + \boldsymbol{q}_r \circ \boldsymbol{p}_j - \boldsymbol{q}_i \circ \boldsymbol{p}_k)\boldsymbol{\mathsf{j}} \nonumber \\
&+& (\boldsymbol{q}_k \circ \boldsymbol{p}_r - \boldsymbol{q}_j \circ \boldsymbol{p}_i + \boldsymbol{q}_i \circ \boldsymbol{p}_j + \boldsymbol{q}_r \circ \boldsymbol{p}_k)\boldsymbol{\mathsf{k}}  
\label{equa:halproduct1}
\end{align}

\noindent where $\circ$ denotes the element-wise product.
We note that the Hamilton product is not commutative, i.e., $q \otimes p \neq p \otimes q$.

We can derived a product of a quaternion matrix $\boldsymbol{W} \in \mathbb{H}^{m \times n}$ and a quaternion vector $\boldsymbol{p} \in \mathbb{H}^{n}$ from Equation~\ref{equa:halproduct_newform} as follow:

\begin{equation}
\boldsymbol{W} \otimes \boldsymbol{p} = 
\begin{bmatrix}
1\\
\boldsymbol{\mathsf{i}}\\
\boldsymbol{\mathsf{j}}\\
\boldsymbol{\mathsf{k}}
\end{bmatrix}^\top
\begin{bmatrix}
\boldsymbol{W}_r & -\boldsymbol{W}_i & -\boldsymbol{W}_j & -\boldsymbol{W}_k\\
\boldsymbol{W}_i & \boldsymbol{W}_r & -\boldsymbol{W}_k & \boldsymbol{W}_j\\
\boldsymbol{W}_j & \boldsymbol{W}_k & \boldsymbol{W}_r & -\boldsymbol{W}_i\\
\boldsymbol{W}_k & -\boldsymbol{W}_j & \boldsymbol{W}_i & \boldsymbol{W}_r
\end{bmatrix}
\begin{bmatrix}
\boldsymbol{p}_r\\
\boldsymbol{p}_i\\
\boldsymbol{p}_j\\
\boldsymbol{p}_k
\end{bmatrix}  
\label{equa:halproduct_newform_matrix}
\end{equation}
where $\boldsymbol{p}_r$, $\boldsymbol{p}_i$, $\boldsymbol{p}_j$, and $\boldsymbol{p}_k \in \mathbb{R}^n$ are real vectors; and $\boldsymbol{W}_r$, $\boldsymbol{W}_i$, $\boldsymbol{W}_j$, and $\boldsymbol{W}_k \in \mathbb{R}^{m \times n}$ are real matrices.

\textbf{Quaternion-inner product.}  The quaternion-inner product $\bullet$ of two quaternion vectors $\boldsymbol{q}$ and $\boldsymbol{p} \in \mathbb{H}^n$ returns a scalar as:

\begin{equation}
\boldsymbol{q} \bullet \boldsymbol{p} = \boldsymbol{q}_{r}^\textsf{T}\boldsymbol{p}_{r} + \boldsymbol{q}_{i}^\textsf{T}\boldsymbol{p}_{i} + \boldsymbol{q}_{j}^\textsf{T}\boldsymbol{p}_{j} + \boldsymbol{q}_{k}^\textsf{T}\boldsymbol{p}_{k} 
\end{equation}

\textbf{Quaternion element-wise product.} 
We further define the element-wise product of two quaternions vector $\boldsymbol{q}$ and $\boldsymbol{p} \in \mathbb{H}^n$ as follow:

\begin{equation}
    \boldsymbol{p} * \boldsymbol{q} = (\boldsymbol{q}_r \circ \boldsymbol{p}_r) + (\boldsymbol{q}_i \circ \boldsymbol{p}_i) \boldsymbol{\mathsf{i}} + (\boldsymbol{q}_j \circ \boldsymbol{p}_j) \boldsymbol{\mathsf{j}} + (\boldsymbol{q}_k \circ \boldsymbol{p}_k) \boldsymbol{\mathsf{k}}  
\end{equation}

\bibliographystyle{splncs04}
\bibliography{references}

\end{document}